\pdfoutput=1
\PassOptionsToPackage{table,xcdraw,dvipsnames}{xcolor}
\documentclass[11pt]{article}

\usepackage[final]{acl}

\usepackage{times}
\usepackage{latexsym}
\usepackage{booktabs}
\usepackage{tabularx}
\usepackage[T1]{fontenc}
\usepackage[utf8]{inputenc}

\usepackage{microtype}

\usepackage{inconsolata}

\usepackage{graphicx}

\title{DayDreamer at CQs-Gen 2025: Generating Critical Questions through Argument Scheme Completion}

\author{Wendi Zhou \and  Ameer Saadat-Yazdi \and Nadin Kökciyan \\
        School of Informatics, \\ University of Edinburgh \\ 
        \{wendi.zhou, ameer.saadat, nadin.kokciyan\}@ed.ac.uk}

\begin{document}
\maketitle
\begin{abstract}
Critical questions are essential resources to provoke critical thinking when encountering an argumentative text. We present our system for the Critical Questions Generation (CQs-Gen) Shared Task at ArgMining 2025 \cite{figueras2025benchmarkingcriticalquestionsgeneration}.  
Our approach leverages large language models (LLMs) with chain-of-thought prompting to generate critical questions guided by Walton's argumentation schemes. 
For each input intervention, we conversationally prompt LLMs to instantiate the corresponding argument scheme template to first obtain structured arguments, and then generate relevant critical questions.
% utilises the given argument scheme for each intervention and prompts the LLM to instantiate the corresponding template to extract structured arguments. 
% When the number of critical questions derived from a scheme is insufficient, the system dynamically generates additional questions by continuing the dialogue context. 
Following this, we rank all the available critical questions by prompting LLMs to select the top 3 most helpful questions based on the original intervention text. This combination of structured argumentation theory and step-by-step reasoning enables the generation of contextually relevant and diverse critical questions. 
Our pipeline achieves competitive performance in the final test set, 
% We demonstrate the effectiveness of our system through both qualitative examples and task-based evaluation, 
showing its potential to foster critical thinking given argumentative text and detect missing or uninformed claims. Code available at \href{https://git.ecdf.ed.ac.uk/s2236454/DayDreamer-CQs-Gen}{DayDreamer}.
\end{abstract}

\section{Introduction}

In this paper, we present a system description for our contribution to the ArgMining 2025 shared task CQs-Gen \cite{figueras2025benchmarkingcriticalquestionsgeneration}. Critical questions are an approach to evaluating arguments by providing criteria upon which an argument can be accepted. The argument can be considered acceptable if all the critical questions are satisfactorily answered~\cite{walton2005nature}. 

In recent years, there has been an increasing interest in developing systems that can automate this process, aiming to improve the efficiency and reliability of argument evaluation. Our approach leverages advanced natural language processing techniques and machine learning algorithms to generate contextually relevant and diverse critical questions.

The system we propose not only identifies key components of an argument but also generates questions that challenge the premises, evidence, and reasoning used in forming conclusions. By doing so, it assists in uncovering potential weaknesses or biases within the argument, thus facilitating more rigorous and comprehensive critical thinking.

Our contribution to the CQs-Gen shared task \cite{figueras2025benchmarkingcriticalquestionsgeneration} is rooted in an approach that integrates argumentation theory with a large-scale language model, allowing our system to understand complex argument structures. Our system relies on the identification of argument schemes according to the taxonomy defined by Walton~\cite{Walton2008Argumentation}.

% This paper will discuss the architectural design of our system, detailing the methods employed in the data pre-processing, question generation, and evaluation phases. We will also present our system's performance results in the shared task, highlighting the strengths and areas for further improvement. Through this research, we aim to advance the field of computational argumentation and contribute valuable insights to the development of automated critical question generation systems.

\section{Background}

In \citet{Walton2008Argumentation}, the authors develop a comprehensive framework of argument schemes from which critical questions can be derived. An argument scheme is a structured pattern of reasoning associated with a common form of argument. These schemes can be used to analyse and evaluate arguments, particularly in everyday discourse where informal logic is often applied. Not only does this work categorise various types of arguments but it also provides critical questions for each scheme that help in assessing arguments. In their work, 26 Argument Schemes are described with associated critical questions. For example, one common scheme is the \textit{Argument from Expert Opinion} shown in Table \ref{tab:scheme-example}.

\begin{table}[h]
\centering
\noindent\begin{tabular}{p{0.2\linewidth} p{0.7\linewidth}}
\toprule
\multicolumn{2}{c}{\textbf{Argument from Expert Opinion}}\\
\midrule
\textit{Premise} & \texttt{E} is an expert in domain \texttt{D}. \\
\textit{Premise} & \texttt{E} asserts that \texttt{A} is true (false). \\
\textit{Conclusion} & \texttt{A} may plausibly be accepted (rejected). \\
\bottomrule
\end{tabular}
\caption{Scheme for \textit{Argument from Expert Opinion}}
\label{tab:scheme-example}
\end{table}

Critical questions are employed to scrutinise and challenge arguments constructed using argument schemes. These questions aim to identify potential weaknesses or gaps in the argument. Each argument scheme has its own set of critical questions. For the \textit{Argument from Expert Opinion}, the critical questions are shown in Table \ref{tab:cqs}.

\begin{table}[h]
\centering
\noindent\begin{tabular}{p{0.9\linewidth}}
\toprule
\textbf{Critical Questions~(CQs)} \\
\midrule
CQ1: {Is \texttt{E} a credible expert in domain \texttt{D}?} \\
CQ2: {Is \texttt{A} consistent with what other experts assert?} \\
CQ3: {Is \texttt{E}'s assertion based on reliable evidence?} \\
CQ4: {Is there any bias or conflict of interest?} \\
CQ5: {Is the argument plausible irrespective of expert opinion?} \\
\bottomrule
\end{tabular}
\caption{Critical Questions associated with the \textit{Argument from Expert Opinion}}
\label{tab:cqs}
\end{table}
These questions guide the evaluator in determining the robustness of the argument by challenging them to assess the credibility of the expert, the quality of the evidence, and any external influences that may affect the truth value of the expert's assertion.

\section{Related Work}
Several works approach the automatic identification of argument schemes as a multiclass classification problem. Starting from raw text, the goal is to label the text according to the scheme of reasoning being used~\cite{ visser_revisiting_2018, rigotti_inference_2019}. Others take this a step further and seek to instantiate the scheme based on the input text~\cite{saadat-yazdi_beyond_2024, jo_classifying_2021, ruiz-dolz_nlas-multi_2024}. The latter approach considers the problem of scheme identification as a two-step process of scheme classification, followed by instantiation, or a direct sequence-to-sequence translation problem. We combine these two approaches by choosing scheme labels that describe the set of schemes we wish to identify first. However, our goal is to automatically find the exact span of text to which a particular scheme applies, as well as the instantiation of the scheme.

Automatic critical question generation is less studied, with \citet{calvo-figueras-agerri-2024-critical} being the only work that explicitly undertakes this investigation. Several other works, however, touch upon aspects of automated question generation in broader contexts. \citet{mulla_automatic_2023} survey a number of approaches ranging from rule-based to neural approaches for automatic question generation, finding that modelling the task as a sequence-to-sequence learning problem seems to be the most promising direction.  

\section{Critical Question Generation Pipeline}\label{sec:cqg_pipeline}

\begin{figure}[h]
    \centering
    \includegraphics[width=\linewidth]{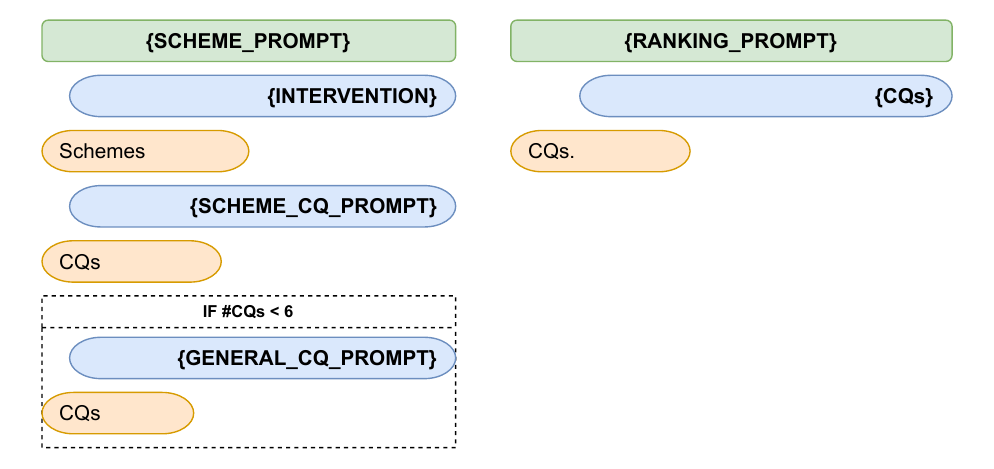}
    \caption{Conversational structure of our approach. The system prompt is shown in \textcolor{PineGreen}{green}, user prompts in \textcolor{RoyalBlue}{blue}, and LLM responses in \textcolor{Peach}{orange}. The text associated with user and system prompts can be found in Appendix~\ref{sec:appendix}.}
    \label{fig:conversation}
\end{figure}

We now introduce the three main stages within our critical question generation pipeline: \textit{Argument Extraction}, \textit{Critical Question Generation} and \textit{Ranking}. Since our pipeline relies on chain-of-thought prompting with LLMs, the output of each stage would be the input for the next one. This conversational structure is depicted in Figure \ref{fig:conversation}. 
\paragraph{Argument Extraction} In this stage, we utilised a comprehensive approach to extract arguments with the intervention text as input. 
Each intervention text was paired with a list of schemes in the provided dataset, which indicates the types of arguments that have been made in the intervention. To utilise this,
% Initially, we identified scheme names within the dataset and the original intervention. With these scheme names at hand, 
we collected the definition of all the argument schemes from \cite{Walton2008Argumentation} and provided them to LLMs for template instantiation~(prompt in Table \ref{tab:prompt-arg-extract}), thereby generating structured arguments. 
% we employed scheme templates to prompt the model, thereby facilitating the extraction of relevant arguments. 
This step provided a structured representation and categorisation of arguments, laying the foundation for critical question generation.

\paragraph{Critical Question Generation} After successfully extracting the arguments, the next phase involved generating critical questions pertinent to each scheme. This was also accomplished by referencing Walton's work~\cite{Walton2008Argumentation}, which provides a well-established framework of critical questions for each scheme. 
With the prompt in Table \ref{tab:prompt-cqg}, we complemented the LLMs ability on critical questions generation with this well-defined framework, providing guidance for generating more relevant and helpful questions by helping the models to hallucinate less.
% We complemented this framework with the capabilities of a Language Learning Model (LLM) to ensure comprehensive and relevant question generation. 
Occasionally, this process would result in fewer than three critical questions. To address this, we introduced one more turn (the dash box in Figure \ref{fig:conversation}) that directly prompts LLMs to generate additional critical questions based on the chatting history when the total number of critical questions is insufficient for the next ranking stage (prompt in Table \ref{tab:prompt-mcq}).
% This dual approach aimed at enriching the critical evaluation process by providing a robust set of questions.
\begin{table*}[t]
\centering
\begin{tabular}{lcccc}
\toprule
\textbf{Method}           & \textbf{Useful}               & \textbf{Unhelpful}            & \textbf{Invalid}             & \textbf{N/A}   \\ \midrule
Baseline     & \cellcolor[HTML]{65BF7D}72.04 & \cellcolor[HTML]{FCFCFF}13.80 & \cellcolor[HTML]{FCFCFF}3.94 & \cellcolor[HTML]{FCFCFF}10.22 \\
Direct Prompting         & \cellcolor[HTML]{FCFCFF}56.81 & \cellcolor[HTML]{CAE8D4}12.19 & \cellcolor[HTML]{81CA95}1.79 & \cellcolor[HTML]{63BE7B}29.21 \\
$Con$              & \cellcolor[HTML]{C0E4CB}62.90 & \cellcolor[HTML]{E6F3EC}13.08 & \cellcolor[HTML]{63BE7B}1.25 & \cellcolor[HTML]{97D4A8}22.76 \\
$Con_{+ss}$     & \cellcolor[HTML]{A7DAB6}65.41 & \cellcolor[HTML]{EBF5F0}13.26 & \cellcolor[HTML]{F1F7F6}3.76 & \cellcolor[HTML]{C1E5CC}17.56 \\
$Con_{+ss+rank}$        & \cellcolor[HTML]{8BCE9D}68.28 & \cellcolor[HTML]{C5E5CF}12.01 & \cellcolor[HTML]{FCFCFF}3.94 & \cellcolor[HTML]{D0EAD9}15.77 \\ 
$Con_{+ss+rank-er}$ & \cellcolor[HTML]{63BE7B}72.22 & \cellcolor[HTML]{63BE7B}8.78  & \cellcolor[HTML]{BEE3CA}2.87 & \cellcolor[HTML]{CDE9D6}16.13 \\
\bottomrule

\end{tabular}
\caption{Ablation study of our model showing how different model choices affect validation performance. All the numbers are the percentage of the number of critical questions with the label. $Con$ is the abbreviation of "Conversational prompting". $Con_{+ss}$ represents that we include "sort scheme" technique on top of the conversational prompting design. Similarly, $Con_{+ss+rank}$ represents that we include prompt tuning for ranking, and $Con_{+ss+rank-er}$ means we remove the scheme templates starting with "ER" as input for LLMs. We use \textbf{N/A} to represent the fourth label in the automated evaluation: "not\_able\_to\_evaluate".\label{tab:val_result}}
\end{table*}
\paragraph{Ranking of Critical Questions} The final stage of our pipeline focused on ranking the generated critical questions. Ranking is done with a new chat history as we are only interested in the original intervention and the generated critical questions. Using the prompt in Table \ref{tab:prompt-cqr}, we present these to LLMs and task them with assessing and ranking the questions based on the helpfulness of the questions. Then, LLMs select the top three most helpful questions as the final output. This ranking process was crucial in choosing the most significant critical questions that would contribute to more in-depth critical thinking, considering the intervention.

\section{Results}

\subsection{Final Evaluation}

We obtained the $4^{th}$ place out of 13 teams that participated, having 60 \textit{Helpful} questions, 25 \textit{Unhelpful} questions and 17 \textit{Invalid} questions. This result comes from our first run result using GPT-4o-mini with manual evaluation.

\begin{figure}[t]
    \centering
    \includegraphics[width=\linewidth]{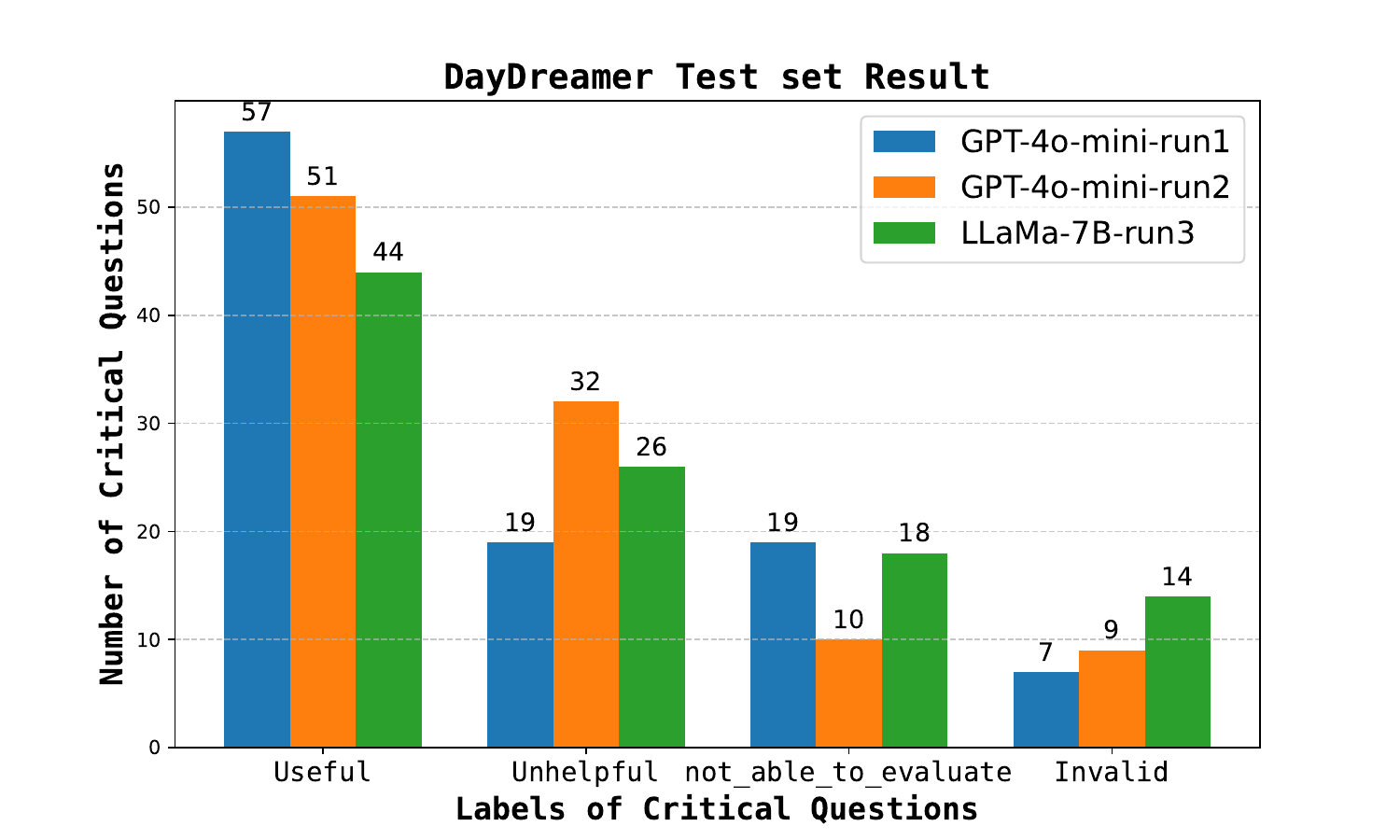}
    \caption{The automated test set evaluation results across three runs. The first two runs are implemented with GPT-4o-mini and the third one is with LLaMa-7B.}
    \label{fig:results}
\end{figure}

Figure \ref{fig:results} shows the comparison of our three submissions, where our critical question generation pipeline is combined with two backbone models: \textit{GPT-4o-mini} from
OpenAI\footnote{\url{https://platform.openai.com/docs/models/gpt-4o-mini}} and \textit{LLaMa-3.1-8B-Instruct} \cite{grattafiori2024llama3herdmodels}. 
% We evaluated our critical question generation pipeline based on GPT-4o-mini and LLAMA-7B. 
Runs $1$ and $2$ use GPT model twice to assess the stability of our results. Overall, \texttt{GPT-4o-mini-run1} achieves the best performance, generating more Helpful critical questions while producing fewer Invalid and Unhelpful ones. \texttt{GPT-4o-mini-run2} shows a similar but slightly worse profile, suggesting some instability in our pipeline. In contrast, \texttt{LLaMa-7B-run3} demonstrated the lowest response quality compared to other runs, with a tendency toward less helpful and more error-prone outputs. These results highlight the 
better capability of GPT-4o models in critical question generation compared to LLaMa-7B; however, our pipeline fails to achieve consistent performance in unlocking their full potential.

% relative robustness and reliability of the GPT-4o variants, particularly the first run, in handling complex and critical questions.

\subsection{Pipeline Optimization on Validation set}
In Table \ref{tab:val_result}, we list all experiment results on the validation set that we conduct to optimize our critical question generation pipeline. Although the baseline method, where we simply prompt the GPT-4o-mini model with the same instruction as \cite{calvo-figueras-agerri-2024-critical}, achieves the highest percentage for \textit{Useful} questions, our optimization goal is to minimize the number of \textit{Invalid} and \textit{Unhelpful} critical questions rather than maximize the number of \textit{Helpful} ones. Focusing solely on having a higher number of \textit{Helpful} questions may lead to overfitting, as 75\% of the questions in the validation set are generated by LLMs. 
% Compared to the baseline method, , our chain-of-thought pipeline using the argumentation scheme results in a lower number of .

We implement our pipeline both with \textit{direct prompting} of the LLMs as well as \textit{conversational prompting}. For direct prompting, we prompt the LLM separately in each stage of our pipeline, which means we take the output of the previous stage and use it together with the instructions of this stage as the input. On the other hand, we prompt LLMs in a conversational manner by keeping a list of chat history messages. In this way, we only provide this stage's instruction and additional helpful information in the prompt because the response of LLMs from the previous stage already exists in the history messages. When comparing the results from $Con$ and \textit{Direct Prompting} (in Table \ref{tab:val_result}), 
% In Table \ref{tab:val_result}, 
we observe a higher percentage of \textit{Useful} critical questions with a similar percentage of \textit{Invalid} and \textit{Unhelpful} ones. Therefore, we build on top of the conversational prompting method to enhance our pipeline.

Each intervention could be related to a long list of scheme names, and we observe that LLMs tend to hallucinate while having more than two scheme templates as input for argument extraction (Section~\ref{sec:cqg_pipeline}).
% Since we extract arguments based on the scheme templates , 
Initially, we feed those schemes into LLMs with a sliding window where the window size is 2. However, the scheme names within the list are not unique, and the same scheme name could occur in different positions. This window size limits LLMs to extract diverse arguments following the same scheme, as LLMs do not remember what arguments have been extracted with this scheme. To generate more diverse arguments and critical questions, we overcome this challenge with the "sort scheme" technique, where we sort the scheme names in the list and provide all the occurrences of the same scheme names to LLMs together. 
This approach enables LLMs to estimate the number of argument instances within the intervention that follow the scheme template, thus extracting them all together. 
% From Table \ref{tab:val_result}
There is an evident increase in the number of \textit{Useful} questions and \textit{Invalid} ones from $Con$ to $Con_{+ss}$ in Table \ref{tab:val_result}, justifying that sorting scheme names could result in more diverse critical question generation. Furthermore, we improve the number of \textit{Helpful} questions by modifying the instructions for the ranking stage.

Since our pipeline involves a chain-of-thought prompting, the response of LLMs for each stage could have a great influence on the next stage. We perform a bad case analysis to correlate the quality of the generated critical questions with the scheme types. Unsurprisingly, we notice that most of the \textit{Invalid} critical questions are generated using the schemes that start with ``ER'' (such as ``ERPracticalReasoning'', ``ERExpertOpinion'', etc), which are not defined in \cite{Walton2008Argumentation}. Since we failed to find the accurate definition, we filled the scheme templates with the corresponding scheme that does not start with ``ER''. For example, we used the scheme content of "PracticalReasoning" for the scheme ``ERPracticalReasoning''. However, this inaccurate scheme definition seems to confuse LLMs from extracting correct arguments from the intervention, thus resulting in poor critical question generation. So, we decide not to provide any template to LLMs for these four schemes and let them generate the critical questions based purely on the intervention text. The difference between results from $Con_{ss+rank}$ and $Con_{+ss+rank-er}$ in Table~\ref{tab:val_result} suggests that LLMs can generate higher quality critical questions without misleading scheme templates. Therefore, the quality of the scheme template has a great impact on our pipeline.

% In this way, LLMs could know how many arguments following this scheme occur in the intervention and extract them all together.
% first implement the pipeline through direct prompting LLMs for each stage, where we are giving the output from the 

\section{Conclusion}
The findings of our study underscore the significant impact that argument schemes have on the critical question generation process. Our analysis indicates that the accurate definition and implementation of schemes are crucial for extracting valid arguments and enhancing the overall effectiveness of the pipeline. Future work may focus on improving the ability of language models to correctly identify schemes and generate appropriate critical questions accordingly. Constructing a compendium of argument scheme definitions used in the dataset, alongside generating critical questions, would also likely improve results in follow-up work, as it would avoid the issues we found with ``ER'' schemes.
~\\~\\
\noindent \textbf{Acknowledgment. }This work was supported by the Edinburgh-Huawei Joint Lab grants CIENG4721 and CIENG8329.
\section*{Limitations}

As discussed in our results, the key limitation of this is the lack of definitions of argument schemes for certain cases. We also found that certain schemes used in the dataset were not provided with critical questions in \citet{Walton2008Argumentation}, preventing us from generating critical questions once the scheme has been extracted.

\bibliography{custom, anthology}
\begin{appendix} 
    
\section{Prompts for LLMs}
\label{sec:appendix}
\begin{table}[h]
    % \centeing
    \footnotesize
    \begin{tabularx}{0.48\textwidth}{|>{\raggedright\arraybackslash}X|}
         \hline
         \\
Extract arguments for each of the scheme in \{\textbf{{scheme\_name}}\} from the input paragraph. 
  These schemes are defined as follows:

  \\
 \{\textbf{{scheme\_description}}\}\\
  If no argument can be extracted to fit the scheme, extract the main arguments with premise and conclusion.
  
  \\
\hline
    \end{tabularx}
    \caption{\textbf{SCHEME\_PROMPT} Prompt for the \textbf{Argument Extraction} stage. \{\textbf{{scheme\_name}}\} is the placeholder for the scheme names paired with this intervention. \{\textbf{{scheme\_description}}\} is the placeholder for the scheme definition in \cite{Walton2008Argumentation}.}
    \label{tab:prompt-arg-extract}
\end{table}

\begin{table}[htb]
    % \centeing
    \footnotesize
    \begin{tabularx}{0.48\textwidth}{|>{\raggedright\arraybackslash}X|}
     \hline
     \\
        \{\textbf{{cq\_template}}\}
    \\
    \\
  With the help of the information above, generate a list of critical questions to ask regarding the extracted arguments.\\
  You may rephrase the critical question to make it more fluent.\\
  Return only a list questions as defined below: \\
  
  [\{"CQ1": "the content of the critical question"\}, ...]

    \\
    \hline
    \end{tabularx}
    \caption{\textbf{SCHEME\_CQ\_PROMPT} Prompt for the \textbf{Critical Question Generation} stage. \{\textbf{{cq\_template}}\} is the placeholder for the defined critical question template related to each scheme.}
    \label{tab:prompt-cqg}
\end{table}

\begin{table}[htb]
    % \centeing
    \footnotesize
    \begin{tabularx}{0.48\textwidth}{|>{\raggedright\arraybackslash}X|}
     \hline
     \\
        \{\textbf{{intervention}}\}
        
    \\
  A helpful critical question can potentially challenge one of the arguments in the text.\\
  Rank and select top three most helpful critical questions.\\
  Return ONLY the question id in a Python list: \\
  ```python \\
  
  [id\_1, ...]

    \\
    \hline
    \end{tabularx}
    \caption{\textbf{GENERAL\_CQ\_PROMPT} Prompt for the \textbf{Ranking of Critical Questions} stage. \{\textbf{{intervention}}\} is the placeholder for the original intervention text.}
    \label{tab:prompt-cqr}
\end{table}

\begin{table}[htb]
    % \centeing
    \footnotesize
    \begin{tabularx}{0.48\textwidth}{|>{\raggedright\arraybackslash}X|}
     \hline
     \\
     
    \{\textbf{{intervention}}\}

    \\
  A helpful critical question can potentially challenge one of the arguments in the text.\\
  Provide me 3 more critical questions that should be asked given the arguments from the text above.\\
  Return only the questions as following format:\\
  
  [\{"CQ1": "the content of the critical question"\}...]

    \\
    \hline
    \end{tabularx}
    \caption{\textbf{RANKING\_PROMPT} Prompt for generating more critical questions when the available critical questions are insufficient for ranking.}
    \label{tab:prompt-mcq}
\end{table}
\end{appendix}

\end{document}